\pdfoutput=1

\documentclass[11pt]{article}

\usepackage[final]{acl}

\usepackage{times}
\usepackage{latexsym}

\usepackage[T1]{fontenc}
\usepackage{amsfonts}
\usepackage{amsmath}
\usepackage{hyperref}
\usepackage{verbatim}
\usepackage{booktabs}
\usepackage{multirow}
\usepackage{xcolor}
\usepackage{subcaption}
\usepackage{float}
\usepackage[utf8]{inputenc}

\usepackage{microtype}

\usepackage{inconsolata}

\usepackage{graphicx}

\title{InCo-DPO: Balancing Distribution Shift and Data Quality for Enhanced Preference Optimization}

\newcommand{\buaa}{\textsuperscript{\S}}
\newcommand{\baai}{\textsuperscript{\textdagger}}

\author{Yunan Wang{\baai\buaa}, Jijie Li{\baai}, Bo-Wen Zhang{\baai},Liangdong Wang{\baai},Guang Liu\textsuperscript{\textdagger\textnormal{\textasteriskcentered}}\\
{\baai}Beijing Academy of Artificial Intelligence\\
{\buaa}Beihang University\\
\texttt{liuguang@baai.ac.cn}
}

\begin{document}
\maketitle
\let\thefootnote\relax\footnotetext{\textsuperscript{\textasteriskcentered}Corresponding authors}
\begin{abstract}
Direct Preference Optimization (DPO) optimizes language models to align with human preferences. Utilizing on-policy samples, generated directly by the policy model, typically results in better performance due to its \emph{distribution consistency} with the model compared to off-policy samples. This paper identifies the \emph{quality} of candidate preference samples as another critical factor. While the quality of on-policy data is inherently constrained by the capabilities of the policy model, off-policy data, which can be derived from diverse sources, offers greater potential for quality despite experiencing distribution shifts. However, current research mostly relies on on-policy data and neglects the value of off-policy data in terms of data quality, due to the challenge posed by distribution shift. In this paper, we propose InCo-DPO, an efficient method for synthesizing preference data by integrating on-policy and off-policy data, allowing dynamic adjustments to balance distribution shifts and data quality, thus finding an optimal trade-off. Consequently, InCo-DPO overcomes the limitations of distribution shifts in off-policy data and the quality constraints of on-policy data. We evaluated InCo-DPO with the Alpaca-Eval 2.0 and Arena-Hard benchmarks. Experimental results demonstrate that our approach not only outperforms both on-policy and off-policy data but also achieves a state-of-the-art win rate of 60.8 on Arena-Hard with the vanilla DPO using Gemma-2 model.
\end{abstract}

\section{Introduction}
Aligning Large Language Models (LLMs) with human preferences significantly improves their performance in terms of helpfulness, correctness, coherence, and safety \citep{gao2024towards,achiam2023gpt}. Among widely adopted alignment methods, Direct Preference Optimization (DPO) \citep{rafailov2024direct} directly aligns the language model with human preferences using preference pairs (i.e., responses $y_{win} > y_{lose}$ for a given prompt $x$). Preference data, as a critical factor in alignment, has garnered significant attention \cite{cui2024ultrafeedback,zhang2024lists,d2024anchored}.

Preference data can be categorized into \emph{on-policy} data generated by the current policy model itself, and \emph{off-policy} data from other models or human annotations. In LLM alignment, \emph{distributional shift} is a key consideration \cite{yang2024self,thangarasa2024self,zhou2024wpo}. Off-policy data often exhibits a distributional shift relative to the current policy model, leading to instability and hindering convergence. In contrast, on-policy data benefits from distribution consistency, often resulting in better performance. However, we find that the \emph{quality} of the candidate preference samples is another critical factor (see Section \ref{sec:offvson}), and the value of off-policy data in terms of quality is often overlooked. Since on-policy data is generated by the current policy model, its quality is inherently limited by the model's own capabilities (especially for smaller-scale models). On the other hand, off-policy data can be derived from diverse sources such as existing supervised fine-tuning datasets, preference datasets and manually constructed dialogue data. Among these sources, responses from state-of-the-art (SOTA) open-source models or human annotations tend to outperform on-policy data in terms of quality.

Consequently, pure on-policy and off-policy data each have distinct advantages and limitations regarding distributional shifts and data quality. However, current research mostly relies on on-policy data and neglects the value of off-policy data in terms of data quality due to the challenge posed by distribution shift \cite{agarwal2024policy,meng2024simpo,tajwar2024preference,yuan2024self,wu2024meta}. To address this gap, we propose InCo-DPO (Figure \ref{fig:main}), an efficient preference data synthesis method that \underline{In}tegrates on-policy and off-policy data based on \underline{Co}ntinuation. InCo-DPO supports dynamic adjustment of the trade-off between distribution shift and data quality, thus overcoming the challenges posed by the distribution shift of off-policy data and quality limitations of on-policy data. We quantify the quality of preference data with the reward provided by a reward model, and the degree of distributional shift with the normalized output probability by the policy model (see Section \ref{sec:preliminary}). Inspired by the observation that \emph{the reward of partial responses often positively correlates with the reward of full responses} \cite{sunfast}, we introduce a simple yet effective continuation-based data synthesis method that significantly improves response reward with only a minor sacrifice in distribution consistency. For a given instruction, we first sample a few tokens with a SOTA open-source model as a prefix and then directly continue the prefix with the target policy model. Through this pipeline, the reward score of the response can be significantly improved compared to on-policy responses even though when the prefix length is less than 10 tokens. To mitigate the cost of reduced distribution consistency, we reduce the sampling temperature, thereby increasing the likelihood of the model outputting the token consistent with its own predictions.

We evaluate the effectiveness of InCo-DPO with Alpaca Eval 2.0 and Arena-Hard benchmarks across multiple open-source models, datasets and DPO variants. Experimental results demonstrate that our method achieves superior performance compared to both on-policy and off-policy data. Using UltraFeedback instruction dataset and Gemma-2 model, InCo-DPO achieves a SOTA win rate of 60.8 on Arena-Hard with vanilla DPO.

In summary, our contributions are as follows:
\begin{itemize}
\setlength{\itemsep}{0pt}
\item \textbf{Identifying data quality as key factor:} While past research limited to the analysis of distribution shift, we uncover data quality as another key factor for preference learning.
\item \textbf{Exploiting the value of off-policy data:} We highlight the long-overlooked value of off-policy data and efficiently leverage its advantages in reward compared to on-policy data.  
\item \textbf{Proposing InCo-DPO:} We introduce a novel preference data synthesis strategy that integrates off-policy and on-policy data through a continuation-based method. InCo-DPO dynamically balances distribution shift and reward, thus significantly improving the alignment performance of DPO compared to pure on-policy and off-policy settings.
\item \textbf{Extensive empirical validation:} We conduct comprehensive experiments on multiple open-source datasets, models, and DPO variants, demonstrating that preference data synthesized via InCo-DPO outperforms both on-policy and off-policy data. 
\end{itemize}

\section{Related Work}

\noindent\textbf{LLM preference data synthesis and augmentation}. High-quality training data are challenging to collect and often limited in quantity. To address this issue, researchers have proposed various methods for synthesizing and augmenting data with LLMs \cite{gao2023self,gandhi2024better,wang2023self,cui2023ada,ye2023generating}. Ultrafeedback employs models of different scales and series to generate diverse responses based on various prompting principles, with preference annotations provided by GPT-4. SDFT \cite{yang2024self} leverages the trained model to rewrite downstream SFT data to mitigate the impact of distribution gap. WPO \cite{zhou2024wpo} and DNO \cite{rosset2024direct} incorporate an off-policy sample from strong model as teacher response to build preference pairs. Unlike this sample-level data augmentation approach, InCo-DPO dynamically adjusts data quality and distribution shift by introducing token-level strong external model response prefixes, achieving more flexible control.

\smallskip
\noindent\textbf{On-policy vs. off-policy}. When applying RL to the post-training of the LLM, on-policy learning \citep{zheng2023secrets,santacroce2023efficient,dong2024rlhf} samples training data from the policy model (i.e., the model under training or to be trained) itself while off-policy learning \citep{meng2024simpo,xiong2024iterative,rosset2024direct} generates samples with other methods (e.g., by other models or humans). Compared to on-policy learning, off-policy learning suffers from the drawback of out-of-distribution training samples being inconsistent with the current model's policy, leading to suboptimal performance. To address this, iterative preference optimization \citep{yuan2024self,wu2024meta} continuously updates the policy model in many iterative with the preference pairs generated and scored by the policy itself at the end of each iteration. However, synthetic samples from the policy model have an upper bound in quality, falling short of those produced by state-of-the-art models, thus limiting the effectiveness of on-policy training for less capable models. To overcome this limitation, we propose InCo-DPO where the policy model generates its own responses using strong models' partial response as a prefix to produce high-reward training samples.

\section{Method}
In this section, we first provide the theoretical background of Direct Preference Optimization (DPO). Then we introduce a weight to measure the consistency of response sequences with the policy model in Section \ref{sec:preliminary}. Next, we explain, based on preliminary experiments, the significance of the reward of responses for training in DPO in Section \ref{sec:offvson}, as well as the correlation between partial responses and overall response rewards in Section \ref{sec:positive}. Finally, we present InCo-DPO in Section \ref{sec:method}, which aims to improve response rewards while maintaining policy consistency.
\subsection{Preliminaries}
\label{sec:preliminary}

\noindent\textbf{DPO}. Reinforcement Learning from Human Feedback (RLHF) consists of two stages: the training of the reward model and the online learning of the policy model. This process is complex, unstable and requires substantial computational resources. DPO treats the policy model as an intrinsic reward model, aligning it directly with preference pairs. This approach eliminates the need for fitting a separate reward model, sampling multiple responses during training or conducting extensive hyperparameter tuning. To achieve this, DPO first establishes a mapping between the reward function $r(x,y)$ and the optimal policy $\pi^*$:
\begin{equation}
  \label{eq:rxy}
  \resizebox{0.8\linewidth}{!}{
  $r(x,y)=\beta\log{\frac{\pi^*(y|x)}{\pi_{ref}(y|x)}}+\beta\log{Z(x)}$},
\end{equation}
where $Z(x)$ represents the partition function and $\beta$ serves as the penalty coefficient for the KL divergence between the output distribution of $\pi^*$ and the reference $\pi_{ref}$ which is normally the supervised fine-tuned model. $\beta$ prevents $\pi^*$ deviating excessively from $\pi_{ref}$, thereby stabilizing the training process.

The training objective of the reward model in RLHF is
characterized by the following equation:
\begin{equation}
  \label{eq:lr}
  \resizebox{0.4\textwidth}{!}{
    $\mathcal{L}_R(r, \mathcal{D}) = -\mathbb{E}_{(x, y_w, y_l)\sim \mathcal{D}} \left[ \log \sigma(r(x, y_w) - r(x, y_l)) \right],$
}
\end{equation}
Applying Eq.~\eqref{eq:rxy} to Eq.~\eqref{eq:lr}, we have the objective of DPO:
\begin{equation*}
\label{eq:ldpo}
\resizebox{0.46\textwidth}{!}{
    $\mathcal{L}_{DPO}(\pi_\theta, \pi_{ref}) = - \mathbb{E}_{(x, y_w, y_l)\sim\mathcal{D}} \left[\log\sigma\left( \beta\log\frac{\pi_\theta(y_w | x)}{\pi_{ref}(y_w | x)} - \beta \log \frac{\pi_\theta(y_l | x)}{\pi_{ref}(y_l | x)} \right) \right].$
}
\end{equation*}
where $y_w$ and $y_l$ denote the chosen and rejected responses under the instruction $x$, respectively.

Since DPO is derived from the RLHF, when training with off-policy data, it faces the same challenge as off-policy RL. That is, the distributional discrepancies between the data and the policy model lead to instability in training and difficulty in convergence. To achieve better alignment, it is typical to use on-policy responses directly from the current policy to construct the preference data, which have high consistency with the policy model.

\smallskip
\noindent\textbf{Consistency weight}. To evaluate the consistency of the given response $y$ with the policy model $\pi$ under instruction $x$, we adopt the consistency weight proposed by WPO \cite{zhou2024wpo}:
\begin{equation*}
\label{eq:weight}
\resizebox{0.40\textwidth}{!}{
    $w(x, y)=\exp\left(\frac{1}{|y|}\sum\limits_{t=1}^{|y|}\log \frac{\pi_\theta(y_t|x, y_{<t})}{\sum_{v\in \mathcal{V}}\pi_\theta(v|x, y_{<t})^2}\right),$}
\end{equation*}

Given an instruction $x$ and the generated response $y<t$, the next token $y_t$ is determined by response $y$. The model's actual output probability distribution of the whole vocabulary $\mathcal{V}$ is given by $\pi_{\theta}(v|x,y<t)$. We then compute $\sum_{v\in \mathcal{V}}\pi_{\theta}(v|x,y<t)^2$ to normalize the output probability of $y_t$.

Finally, by averaging the normalized output probabilities for each $y_t$ in $y$, we obtain a weight $w(x,y)$  ranging from 0 to approximately 1.0, which represents the normalized output probability of $y$ by the policy $\pi$. A weight $w(x,y)$ closer to 1.0 indicates a higher consistency between $y$ and $\pi$. The weight of on-policy data is usually above 0.9, while off-policy data is below 0.6.

\subsection{When off-policy data outperforms on-policy data}
\label{sec:offvson}
To elucidate the importance of reward of preference data for training, we present the preliminary observations based on mistral-7b-sft-beta \cite{tunstall2023zephyr}, which we refer as Mistral-Base. Mistral-Base is a fine-tuned version of the pre-trained model Mistral-7B-v0.1, which doesn't undergo sufficient post-training, resulting in less impact of distribution shifts on DPO alignment effectiveness. In this context, the reward of the preference data plays a major role.

For each instruction from Ultrafeedback \cite{cui2024ultrafeedback}, we collect five responses from Mistral-Base. We score the responses with the reward model ArmoRM \cite{wang2024interpretable}. From this, we extracted the responses with the maximum and minimum rewards, forming on-policy preference pairs. Similarly, we gather off-policy preference pairs generated by the significantly stronger model Qwen2.5-72B-Instruct \cite{qwen2.5}, which exhibit markedly higher reward than those produced by Mistral-Base.

\begin{figure}[t]
\centering
\includegraphics[width=0.7\linewidth]{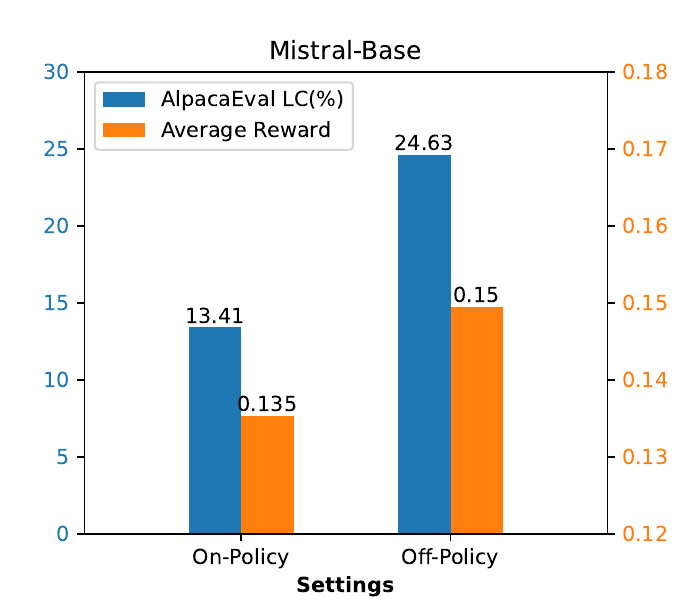}
\caption{Comparison of reward and evaluation results on AlpacaEval2 between on-policy data and off-policy data. We use Mistral-Base as its minimal post-training reduces distribution shift problems, better revealing the relation between reward and performance.}
\label{fig:offvson}
\vspace{-10pt}
\end{figure}

We employ DPO to train Mistral-Base under the same conditions, using these on-policy and off-policy preference pairs. We then obtain evaluation results on Alpaca Eval 2.0. As shown in Figure \ref{fig:offvson}, the Len-Control win rate (LC) of off-policy training is significantly superior to the LC of on-policy training, challenging the conventional belief that training with on-policy data always yields better results than off-policy data. Given that the average reward of off-policy responses generated by Qwen2.5 is considerably higher than that of on-policy responses, we speculate that it is the improvement of reward that significantly enhances the performance.

However, when we train off-the-shelf instruction-tuned models such as Llama-3-8B-Instruct and Gemma-2-9B-it with DPO, they still suffer from distribution shifts, resulting in training performance on off-policy data inferior to on-policy data. We will introduce our method to address this problem in Section \ref{sec:method}.

\subsection{Positive correlation between partial and final response rewards}
\label{sec:positive}
In this section, we present preliminary observations based on Ultrafeedback and Llama-3-8B-Instruct to illustrate the positive correlation between the rewards of partial responses and final responses. For each instruction in Ultrafeedback, we sample five responses from Llama along with their corresponding reward given by ArmoRM. For each response, we extracted the first 100 tokens to form partial responses, which are then scored with ArmoRM. Finally, we analyze the relationship between the reward values of partial and final responses.

\begin{figure}[t]
\centering
\includegraphics[width=0.8\linewidth]{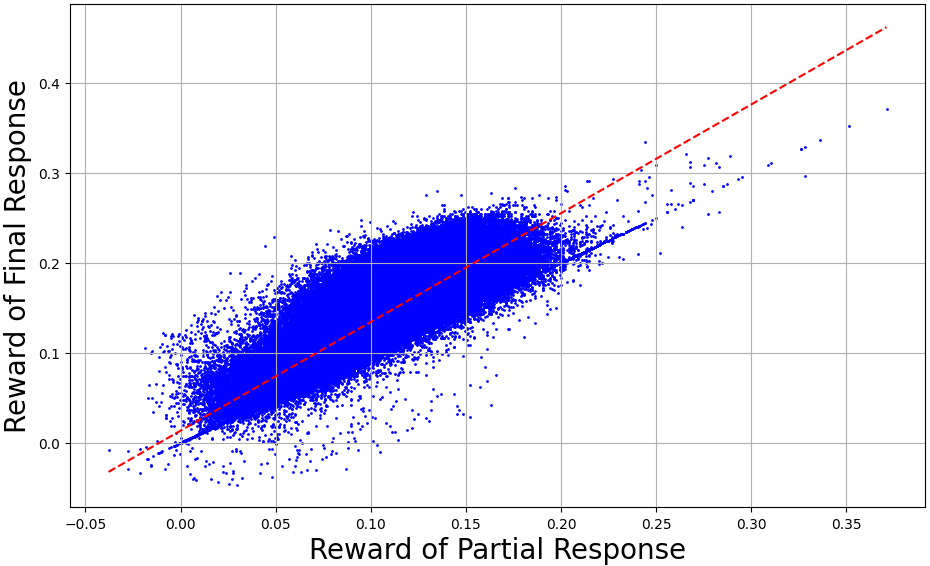}
\caption{Relationship between rewards of partial and final responses with a correlation coefficient of \textbf{0.81}. The red line represents the fitted linear regression line, indicating a significant positive relationship. }
\label{fig:dot}
\vspace{-10pt}
\end{figure}

As shown in Figure \ref{fig:dot}, we can observe a clear positive correlation between the rewards of partial and final responses with a correlation coefficient of 0.81. That is, higher rewards of partial responses are associated with higher rewards of final responses. Therefore, we hypothesize that with partial responses generated by a stronger external model, the response completed by the policy model should have a higher reward than those generated entirely by it. We will verify this hypothesis in Section \ref{sec:prefix}. Notably, some final responses have fewer than 100 tokens. Therefore, the partial responses and the final responses for this portion are identical, resulting in the same reward and forming a straight line under the red line.

\subsection{InCo-DPO}
\label{sec:method}
Based on the observations presented above, this section outlines the synthesis methods for preference data derived from continuation and rewriting, both of which significantly enhance response rewards. We find that rewriting can lead to more substantial losses in policy consistency compared to continuation. Therefore, this paper focuses on the continuation-based method (InCo-DPO). We will compare the experimental results of continuation and rewriting in Section \ref{sec:rewriting}. We provide examples of responses generated by InCo-DPO in Appendix \ref{sec:example}.

\begin{figure*}[t]
\centering
\includegraphics[width=\linewidth]{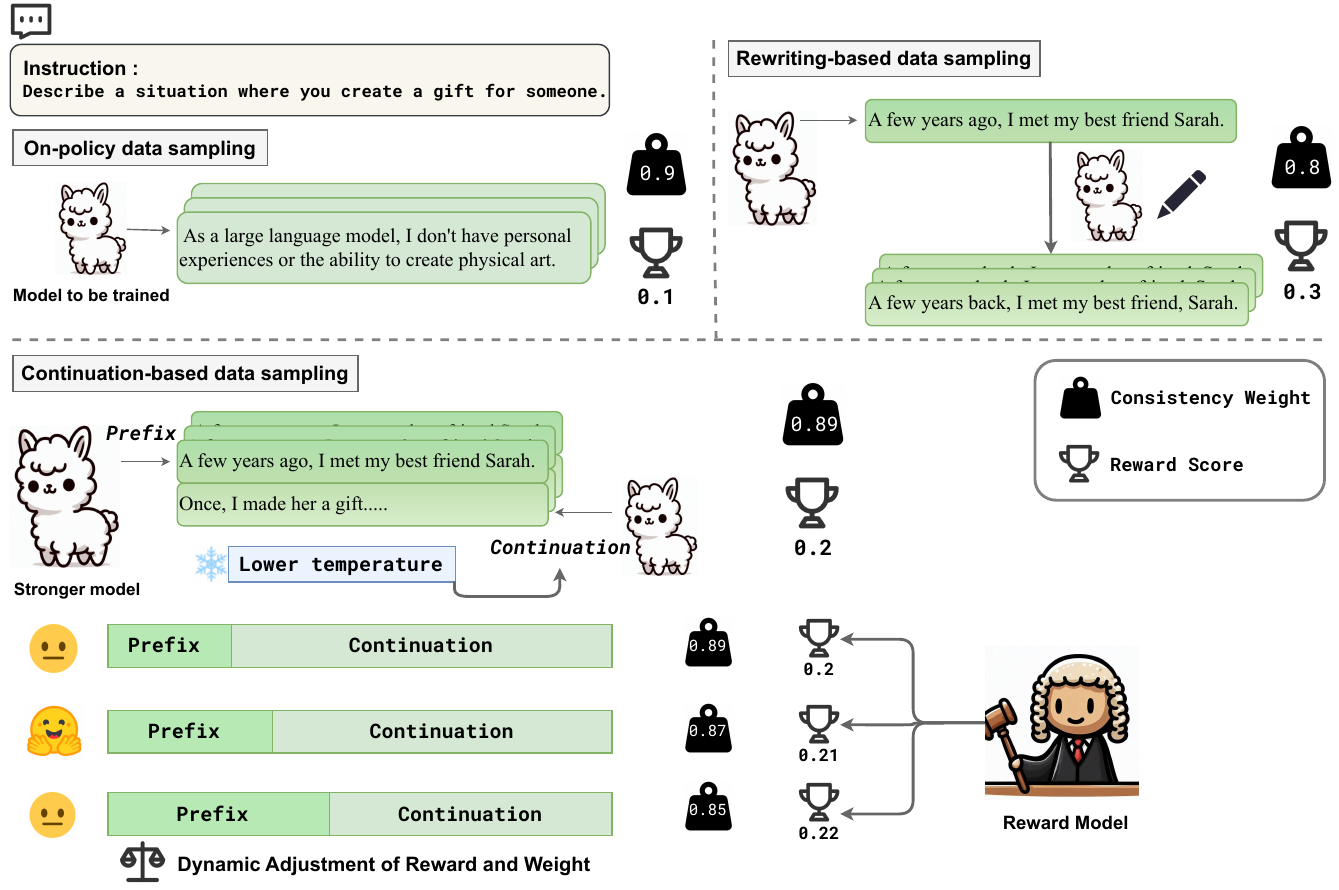}
\caption{Different data synthesis methods. On-policy data is completely sampled from the policy model. Continuation-based sampling (InCo-DPO): the policy model directly continues and completes partial outputs as prefixes from a stronger model under a lower temperature. Rewriting-based sampling: we prompt the policy model to rewrite completed responses from a stronger model.}
\label{fig:main}
\vspace{-15pt}
\end{figure*}

\smallskip
\noindent\textbf{Continuation-based sampling (InCo-DPO).} As illustrated in Figure \ref{fig:main}, for an instruction, different from on-policy sampling generating responses solely with the policy model, we introduce the off-policy part from a stronger model as the prefix $y_p$. The policy model then continues from $y_p$ and completes the response until the stop token is reached. Ultimately, the entire sample consists of $y_p$ from the strong model and the continuation suffix from the policy model. Notably, the continuation method allows for dynamic adjustments to prefix length $|y_p|$, thereby maintaining a balance between reward improvement and consistency loss, which helps identify the optimal trade-off settings. To compensate for consistency loss, we set a lower temperature for sampling. After the sampling, we select the chosen and rejected responses based on scores given by the reward model to construct preference pairs.

\noindent\textbf{Rewriting-based sampling.} Given an instruction $x$, we have responses $y_{ref}$ from a stronger model, which have higher rewards than responses from the policy model. We then construct a prompt with $x$ and $y_{ref}$ as reference. Finally, we prompt the policy model to rewrite $y_{ref}$ in its own tone, thus making $y_{ref}$ align more closely with the model's policy. The prompt can be found in Table \ref{tab:prompt}. However, the effectiveness of this prompt-based approach is heavily dependent on the specific prompt. Compared to continuation, rewriting tends to result in greater losses in policy consistency and lacks the capacity for dynamic adjustments of reward and consistency. Thus, considering the advantages of the continuation method, this paper primarily employs continuation-based sampling.

\section{Experiment}
In this section, we first outline the relevant experimental configurations, including the models, datasets, and times of sampling in Section \ref{sec:setting}. We then present a comparison of InCo-DPO against other sampling methods as the main results in Section \ref{sec:main}. In the ablation study, we emphasize the advantages of InCo-DPO over purely on-policy and off-policy data across different datasets and external models. Then we study the impact of the number of tokens in the continuation prefix and temperature on consistency weight, reward, and overall performance in Section \ref{sec:prefix}. Finally, we compare the performance of InCo-DPO and rewriting in Section \ref{sec:rewriting}. Moreover, we discuss the impact of sampling times on InCo-DPO in Appendix \ref{sec:sampling} and validate the effectiveness of InCo-DPO on tiny models in Appendix \ref{sec:tiny}.

\subsection{Experimental Settings}
\label{sec:setting}

\noindent\textbf{Models and datasets}. We utilize the latest off-the-shelf models, Gemma-2-9B-it \cite{team2024gemma} and Llama-3-8B-Instruct \cite{llama3modelcard}, for main experiments. We utilize Gemma-2 for other experiments. Most experiments are conducted with Ultrafeedback \cite{cui2024ultrafeedback}. For convenience, we use gemma2-ultrafeedback-armorm\footnote{\url{https://huggingface.co/datasets/princeton-nlp/gemma2-ultrafeedback-armorm}} and llama3-ultrafeedback-armorm\footnote{\url{https://huggingface.co/datasets/princeton-nlp/llama3-ultrafeedback-armorm}} for Gemma-2 and Llama-3 respectively, which already contain on-policy samples. We also incorporate Infinity-Preference\footnote{\url{https://huggingface.co/datasets/BAAI/Infinity-Preference}} and HelpSteer2 \cite{wang2024helpsteer2} to demonstrate the generalizability across datasets. For each instruction, we sample five responses under various configurations, including on-policy, off-policy, and InCo-DPO. These responses are scored with the ArmoRM. Then we select the highest and lowest reward responses to construct preference pairs. We use Qwen2.5-72B-Instruct \cite{qwen2.5} as the main external strong model introduced in the off-policy configuration and InCo-DPO. We also include Athene-70B \cite{Athene2024} and Llama3.3-70B \footnote{\url{https://huggingface.co/meta-llama/Llama-3.3-70B-Instruct}} to validate the generalizability across different external models.

\smallskip
\noindent\textbf{Prefix token count and sampling temperature}. As discussed in Section \ref{sec:prefix}, we find a minimal number of prefix tokens can increase the reward. So we perform a grid search of prefix token count over 2, 4 and 8 to find the best balance between reward gain and consistency loss. Specific prefix length can be found in Appendix \ref{sec:implementation}. To mitigate the consistency loss due to the inclusion of off-policy components, we sampled at a lower temperature of 0.6, compared to the on-policy temperature of 0.8.

\smallskip
\noindent\textbf{Evaluation benchmarks}. We utilize two widely adopted evaluation benchmarks in the community: AlpacaEval2 and Arena-Hard, which contain 805 test instructions and 500 challenging prompts respectively. Both AlpacaEval2 and Arena-Hard employ GPT-4 Turbo as the LLM-as-a-judge, comparing the responses of the model under test with those from GPT-4-1106-preview and GPT-4-0314 respectively. We report the length-controlled win rate (LC) for AlpacaEval2 and the win rate (WR) for Arena-Hard.

\smallskip
\noindent\textbf{Baselines and DPO variants}. We adopt DPO as the primary learning objective, comparing InCo-DPO against on-policy setting. We also conduct experiments with WPO \cite{zhou2024wpo} and SimPO \cite{meng2024simpo} to demonstrate the generalizability of InCo-DPO across different variants. Notably, For WPO, we include its results under optimal settings as hybrid-policy (i.e., incorporating one strong model off-policy sample as teacher response apart from four on-policy samples) for comparison. Additionally, for reference, we include results produced by \cite{meng2024simpo} and \cite{wu2024alpha} for IPO \cite{azar2024general}, CPO \cite{xucontrastive}, KTO \cite{ethayarajh2024kto}, ORPO \cite{hong2024orpo} and R-DPO \cite{park2024disentangling} on on-policy data. Further training details can be found in Appendix \ref{sec:implementation}.

\subsection{Main Results and Ablation}
\label{sec:main}

\begin{table*}[!t]
    \centering
    {
    \scalebox{0.63}{
    \begin{tabular}{llcccccccccc}
    \toprule
     \multirow{3}{*}{\textbf{Data Contruction}} & \multirow{3}{*}{\textbf{Objective}} & \multicolumn{5}{c}{\textbf{Llama-3-Instruct (8B)}} & \multicolumn{5}{c}{\textbf{Gemma-2-Instruct (9B)}} \\
     \cmidrule(lr){3-7} \cmidrule(lr){8-12}
     & &\multicolumn{2}{c}{\textbf{Data}}&\multicolumn{3}{c}{\textbf{Benchmark}}&\multicolumn{2}{c}{\textbf{Data}}&\multicolumn{3}{c}{\textbf{Benchmark}}\\
     \cmidrule(lr){3-4}\cmidrule(lr){5-7}\cmidrule(lr){8-9}\cmidrule(lr){10-12}
     & & \multicolumn{1}{c}{\textbf{Reward}}&\multicolumn{1}{c}{\textbf{Weight}}&\multicolumn{1}{c}{\textbf{AE LC(\%)}}&\multicolumn{1}{c}{\textbf{AH WR(\%)}}&\multicolumn{1}{c}{\textbf{Average}}&\multicolumn{1}{c}{\textbf{Reward}}&\multicolumn{1}{c}{\textbf{Weight}}&\multicolumn{1}{c}{\textbf{AE LC(\%)}}&\multicolumn{1}{c}{\textbf{AH WR(\%)}}&\multicolumn{1}{c}{\textbf{Average}} \\
     \midrule
     -&SFT&-&-&22.6&21.2&21.9&-&-&47.9&41.1&44.5 \\
     \midrule
      Off-Policy &DPO&0.151&0.60&46.9&35.1&41.0&0.151&0.54&65.8&54.6&60.2\\
     \midrule
     &DPO& & &48.1&35.2&41.7& & &68.4&56.8&62.6 \\
     &IPO& & &46.8&\underline{36.6}&41.7& & & 62.6&53.5&58.1\\
     &CPO& & &34.1&30.9&32.5& & &56.4&55.2&55.8\\
     On-Policy&KTO&0.144&1.01&34.1&27.3&30.7&0.147&0.91&55.5&53.8&54.7\\
     &ORPO& & &38.1&28.2&33.2& & &56.2&46.2&51.2\\
     &R-DPO& & &48.0&35.1&41.6& & &68.3&57.9&63.1\\
     &SimPO& & &\underline{50.8}&33.0&41.9& & &71.4&59.2&65.3\\
     \midrule
     Hybrid-Policy&WPO&0.145& 0.89 &36.1&16.4&26.3&0.148&0.80&72.3&59.8&66.1\\
     \midrule
    &DPO& 0.146 &0.89 &50.6&\textbf{37.9}&\textbf{44.3}& 0.150 & 0.90 &71.9&60.8&66.4\\
    Ours&SimPO& 0.145 & 0.91 &\textbf{51.0}&33.6&\underline{42.3}& 0.150&0.90  &\underline{72.3}&\underline{62.5}&\underline{67.4}\\
     &WPO& 0.145 & 0.91 &37.4&26.7&32.1&0.149 & 0.92 &\textbf{73.1}&\textbf{63.7}&\textbf{68.4}\\
     \bottomrule
\end{tabular}
}
    }
    \caption{Main results on Alpaca Eval (AE) 2.0 and Arena-Hard (AH). We use UltraFeedback as intruction dataset and Qwen2.5-72B as external model.We highlight the best results in bold and the second-best underlined. InCo-DPO achieves best trade-off between reward and weight and consistently yields better results compared to other data construction strategies under the same objective.}
    \label{tab:main_result}
    \vspace{-10pt}
\end{table*}

\begin{table}[h]
    \centering
    \scalebox{0.69}{
    \begin{tabular}{cccc}
        \toprule
        \textbf{Data} & \multirow{2}{*}{\textbf{Dataset}} & \multirow{2}{*}{\textbf{External Model}} & \multicolumn{1}{c}{\textbf{AE}} \\ 
        \textbf{Construction} &  &  & \multicolumn{1}{c}{\textbf{LC (\%)}} \\ 
        \midrule
        Off-Policy &UltraFeedback &Qwen2.5-72B & 65.8 \\ 
        On-Policy &UltraFeedback &- & 68.4 \\ 
        Ours &UltraFeedback& Qwen2.5-72B& \textbf{71.9} \\
        \midrule
        Off-Policy &Infinity-Preference &Qwen2.5-72B & 67.4 \\ 
        On-Policy &Infinity-Preference &- & 67.7 \\ 
        Ours &Infinity-Preference& Qwen2.5-72B& \textbf{72.7} \\
        \midrule
        Off-Policy &Helpsteer2 &Qwen2.5-72B & 53.0 \\
        On-Policy &Helpsteer2& - & 56.7\\
        Ours &Helpsteer2&Qwen2.5-72B & \textbf{64.0} \\
    \bottomrule
    \end{tabular}}
    \caption{Results of Gemma-2 trained on different data construction strategies across different datasets.}
    \label{tab:generalization_datasets}
    \vspace{-10pt}
\end{table}

\begin{table}[h]
    \centering
    \scalebox{0.73}{
    \begin{tabular}{cccc}
        \toprule
        \textbf{Data} & \multirow{2}{*}{\textbf{Dataset}} & \multirow{2}{*}{\textbf{External Model}} & \multicolumn{1}{c}{\textbf{AE}} \\ 
        \textbf{Construction} &  &  & \multicolumn{1}{c}{\textbf{LC (\%)}} \\ 
        \midrule
        Off-Policy &UltraFeedback &Qwen2.5-72B & 65.8 \\ 
        On-Policy &UltraFeedback &- & 68.4 \\ 
        Ours &UltraFeedback& Qwen2.5-72B& \textbf{71.9} \\
        \midrule
        Off-Policy &UltraFeedback &Athene-70B & 57.2 \\
        On-Policy &UltraFeedback &-& 68.4 \\
        Ours & UltraFeedback &Athene-70B & \textbf{70.1} \\
        \midrule
        Off-Policy &UltraFeedback &Llama3.3-70B & 65.9 \\
        On-Policy &UltraFeedback& - & 68.4\\
        Ours &UltraFeedback&Llama3.3-70B & \textbf{70.4} \\
    \bottomrule
    \end{tabular}}
    \caption{Results of Gemma-2 trained on different data construction strategies across different external models.}
    \label{tab:generalization_external}
    \vspace{-10pt}
\end{table}

\noindent\textbf{InCo-DPO consistently outperforms other data construction strategies or policies, especially significant when adopting vanilla DPO.} As shown in Table \ref{tab:main_result}, InCo-DPO achieves significant improvement in reward and minimal loss in consistency weight and consistently outperforms off-policy, on-policy, and hybrid-policy in terms of Alpaca Eval 2.0 LC and Arena-Hard WR scores when employing different variants of DPO. Notably, when using vanilla DPO, our method achieves a significant improvement of 2.5 to 4 points over the results obtained with on-policy.

\smallskip
\noindent\textbf{The strategy of combining on-policy and off-policy samples is not always applicable with WPO.} WPO improves the performance of off-policy training by multiplying the DPO objective by the consistency weight introduced in Section \ref{sec:preliminary} \cite{zhou2024wpo}. WPO's experiments demonstrate that hybrid-policy yields better performance than pure on-policy data with WPO. However, as shown in Table \ref{tab:main_result}, when Qwen2.5 is selected as the teacher model, the results of Llama-3 with WPO are suboptimal, even underperforming compared to DPO with on-policy data. We hypothesize that this may be attributed to Llama-3's tendency towards catastrophic forgetting with WPO.

\smallskip
\noindent\textbf{Generalization of InCo-DPO.} The main results have demonstrated the generalization of InCo-DPO across different models and DPO variants. We further illustrate the generalization of InCo-DPO on various instruction datasets and external strong models. As shown in Table \ref{tab:generalization_datasets} and Table \ref{tab:generalization_external}, under different configurations, the performance of InCo-DPO consistently surpasses that of purely off-policy and on-policy data. Notably, on-policy performance is always better than off-policy results, indicating that the direct incorporation of off-policy data is inefficient. Furthermore, we observe a significant disparity of up to 12 points between the results of on-policy data and off-policy data generated by Athene-70B, highlighting the instability of off-policy training. However, even when utilizing Athene-70B, InCo-DPO achieves superior performance compared to on-policy data.

\subsection{The impact of prefix token number and sampling temperature on performance}
\label{sec:prefix}
\begin{figure}[t]
\centering
\includegraphics[width=0.8\linewidth]{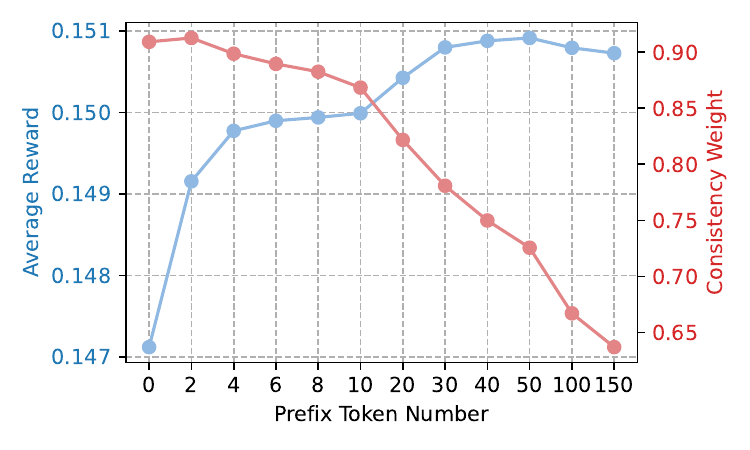}
\caption{The impact of prefix token number on consistency weight and reward. When the Prefix Token Number is less than 10, the consistency weight decreases only slightly while the reward value of the response increases significantly, resulting in an improvement in the final training performance.}
\label{fig:token_weight_reward}
\end{figure}

\begin{figure}[t]
\centering
\includegraphics[width=0.7\linewidth]{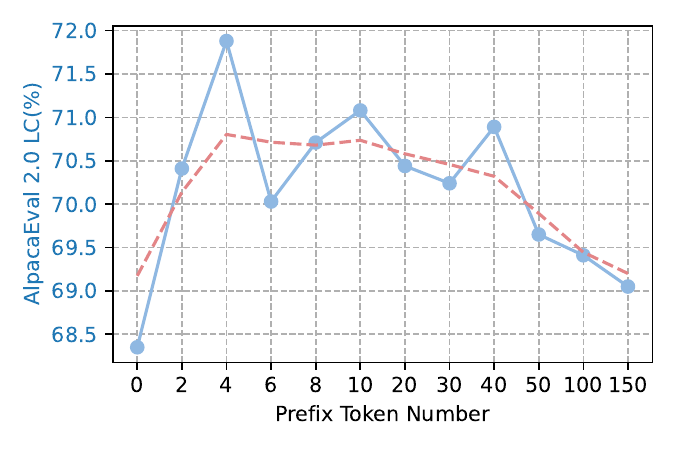}
\caption{The impact of prefix token number on final performance, where the red line represents a Gaussian smoothing curve with a coefficient of 1.0.}
\label{fig:token_eval}
\end{figure}
In this section, we examine the effects of the number of prefix tokens $|y_p|$ from the strong model and the sampling temperature of the policy model on reward, policy consistency, and final performance. The default prefix length is 4 and the default sampling temperature is 0.6.

\smallskip
\noindent\textbf{Impact of prefix length on reward and consistency weight.} As shown in Figure \ref{fig:token_weight_reward}, as prefix length increases, the average reward of the responses rises until it begins to decline when $|y_p|>50$. Notably, when $|y_p| < 10$, reward improves significantly while improvement is relatively modest when $|y_p| > 10$. Conversely, the consistency weight decreases as prefix length increases.

\smallskip
\noindent\textbf{A minimal number of prefix tokens can lead to optimal performance.} According to the Gaussian smoothed curve in Figure \ref{fig:token_eval}, the LC scores on AlpacaEval 2.0 initially rise and then fall as prefix length increases. Optimal performance is observed when $|y_p| = 2, 4, 8, 10$. At these values, the reward shows a significant enhancement compared to the on-policy (i.e., $|y_p| = 0$), while the consistency weights exhibit only a slight decline relative to on-policy weights.

\begin{table}[h]
    \centering
    \scalebox{0.75}{
    \begin{tabular}{ccccc}
        \toprule
        \multirow{2}{*}{\textbf{Temperature}}&\textbf{Consistency}&\multirow{2}{*}{\textbf{Reward}}&\textbf{Reward}&\textbf{AE} \\
         &\textbf{weight}&&\textbf{margin}&\textbf{LC(\%)} \\
         \midrule
         0.5&0.91&0.1498&0.0148&70.9 \\
         0.6&0.90&0.1498&0.0153&71.9 \\
         0.7&0.88&0.1498&0.0157&72.4 \\
         0.8&0.86&0.1497&0.0162&70.6 \\
    \bottomrule
    \end{tabular}}
    \caption{Impact of sampling temperature of policy model on performance.}
    \label{tab:temperature}
    \vspace{-10pt}
\end{table}

\smallskip
\noindent\textbf{Impact of sampling temperature.} As shown in Table \ref{tab:temperature}, the temperature has a minimal effect on the average reward. However, as the sampling temperature increases, the randomness during model decoding rises, leading to a higher probability of outputting tokens that do not align with the current policy. This results in a continuous decline in policy consistency. Nonetheless, this randomness also enhances the diversity of the sampled responses, as reflected in the increase of the reward margin (i.e., $ r(x, y_{w}) - r(x, y_{l})$). Both excessively high and low temperatures can lead to suboptimal performance. When the temperature is too low, it becomes challenging to generate sufficiently distinct responses, resulting in a low reward margin and underfitting in training. Conversely, when the temperature is too high, the reduced policy consistency leads to a decline in performance.

\subsection{Comparison of continuation and rewriting}
\label{sec:rewriting}

We train Gemma-2 on UltraFeedback with Qwen2.5 as external model. We use the prompt specified in Table \ref{tab:prompt} for rewriting. The rest experimental setup remains the same as Section \ref{sec:setting}.

\begin{table}[h]
    \centering
    \setlength{\tabcolsep}{16pt}
    \scalebox{0.77}{
    \begin{tabular}{lccc}
    \toprule
        \textbf{Method}& \textbf{AE LC(\%)} & \textbf{AH WR(\%)}\\
        \midrule
        Continuation$_\textsc{DPO}$ & 71.9 & 60.8 \\
        Continuation$_\textsc{WPO}$ & 73.1 & \textbf{63.7} \\
        Rewriting$_\textsc{DPO}$ & 68.4 & 56.9 \\
        Rewriting$_\textsc{WPO}$ & \textbf{75.7} & 61.0 \\
        \bottomrule
    \end{tabular}}
    \caption{Comparison of performance between continuation and rewriting.}
    \label{tab:rewriting}
    \vspace{-10pt}
\end{table}

\begin{table}[h]
    \centering
    \scalebox{0.73}{
    \begin{tabular}{lcccc}
    \toprule
        \textbf{Attribute}&\textbf{On-Policy}&\textbf{Off-Policy}&\textbf{Continuation}&\textbf{Rewriting} \\
        \midrule
        Reward&0.147&0.151&0.150&0.151\\
        Weight&0.91&0.54&0.90&0.68\\
        \bottomrule
    \end{tabular}}
    \caption{Comparison on reward and consistency weight across data construction strategy.}
    \label{tab:rewriting_reward}
    \vspace{-10pt}
\end{table}

\noindent\textbf{Rewriting demonstrates an advantage in reward values over continuation but is still significantly affected by policy consistency loss}. As shown in Table \ref{tab:rewriting}, when adopting DPO, rewriting underperforms continuation across various benchmarks. In contrast, when WPO is employed to mitigate the impact of policy consistency loss, rewriting shows substantial improvement, surpassing the performance of continuation on AE LC(\%). We further investigate the reward values and consistency weights of both continuation and rewriting responses. As presented in Table \ref{tab:rewriting_reward}, rewriting achieves a higher increase in reward values compared to continuation, but its consistency weights are significantly lower, resulting in poor performance under DPO.

\section{Conclusion}
This paper addresses the balance between two key elements of preference data: data quality and distribution shift. We leverage the quality advantages of off-policy data and the distribution consistency benefits of on-policy data to propose InCo-DPO that integrates both types of data. InCo-DPO significantly enhances reward values while minimizing losses associated with distribution consistency. We conduct extensive experiments demonstrating that InCo-DPO outperforms both pure off-policy and on-policy data. Furthermore, we validated the generalizability of our method across various datasets, models, and DPO variants.

\section{Limitations}
The main limitation of this study lies in the primary dataset UltraFeedback used for experiments, which focuses on the helpfulness, truthfulness, instruction-following, and honesty of responses but lacks alignment regarding safety. Additionally, the subjective evaluation benchmarks, Alpaca Eval2 and Arena-Hard, do not examine preferences for multi-turn instructions. Future work should investigate our approach within a more comprehensive dataset and evaluation framework to ensure the model's safety and evaluate its overall capabilities.
\bibliography{custom}
\appendix
\section{Implementation details}
\label{sec:implementation}

\smallskip
\noindent\textbf{Hyperparameter for InCo-DPO.} Hyperparameter tuning is crucial for DPO and its variants. To ensure a fair comparison, for on-policy setting, we adopt the hyperparameters recommended by \cite{meng2024simpo}. For InCo-DPO, we conduct a hyperparameter search as specified in SimPO \cite{meng2024simpo}. The search range of learning rate is [3e-7, 5e-7, 8e-7, 1e-6]. Our final hyperparameters and prefix length are specified in Table \ref{tab:hyperparameters}.

\smallskip
\noindent\textbf{Other training details.} The following hyperparameter is adopted across all methods. We adopt a batch size of 128. The maximum prompt length is 1800. The maximum sequence length is 2048. All models are trained for a single epoch with a cosine learning rate schedule and a 10\% warmup ratio. The optimizer is AdamW \cite{kingma2014adam}. 

\begin{table*}[h]
    \centering
    \scalebox{0.85}{
    \begin{tabular}{llcccc}
    \toprule
        \textbf{Model} & \textbf{Objective} & \textbf{Learning rate} & $\beta$ & $\frac{\gamma}{\beta}$ & Prefix length\\
        \midrule
         & DPO & 5e-7 & 0.01 & - & 8\\
        Llama-3 & SimPO & 8e-7 & 13 & 0.2 & 2\\
        & WPO & 3e-7 & 0.02 & - & 2 \\
        \midrule
        & DPO & 1e-6 & 0.01 & - & 4\\
        Gemma-2 & SimPO & 1e-6 & 10 & 0.7 & 4\\
        & WPO & 1e-6 & 0.01 & - & 2 \\
        \bottomrule
    \end{tabular}}
    \caption{Hyperparameters for InCo-DPO.}
    \label{tab:hyperparameters}
    \vspace{-10pt}
\end{table*}

\section{The impact of sampling times}
\label{sec:sampling}
\begin{figure}[t]
\centering
\includegraphics[width=0.7\linewidth]{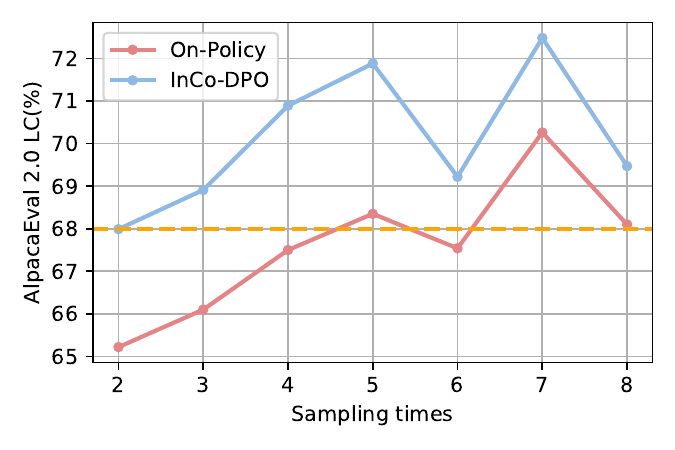}
\caption{The impact of sampling times.}
\label{fig:sample_line}
\vspace{-10pt}
\end{figure}

\smallskip
\noindent\textbf{The impact of sampling times on performance.} We examine the relationship between sampling times and final performance. As shown in Figure \ref{fig:sample_line}, performance increases with sampling times initially and finally stabilizes. The optimal balance between resource consumption and performance is achieved in five samples. Furthermore, performance of InCo-DPO consistently outperforms that of on-policy for the same sampling times. Notably, InCo-DPO requires only two samples to achieve performance comparable to that obtained from five on-policy sampling times.

\section{Prompt for rewriting}
\begin{table*}[h]
    \centering
    \scalebox{1}{
    \begin{tabular}{p{0.9\linewidth}}
        \toprule
        
        You have been provided with the reference response from a strong model to the latest user query. Your task is to \textbf{reflect on} the reference answer, try your best to improve the quality of the answer with your own knowledge and thinking, and finally provide your own response to the question. It is crucial to critically evaluate the information in these responses, recognizing that some may be biased or incorrect. Your response should not simply replicate the given answer but should offer a refined, accurate, and comprehensive reply to the instruction. Ensure your response is well-structured, coherent, and adheres to the highest standards of accuracy, reliability, and helpfulness.

Below is the question and the corresponding reference response.

<start\_of\_question>

<QUESTION\_HERE>

<end\_of\_question>

<start\_of\_reference>

<REFERENCE\_ANSWER\_HERE>

<end\_of\_reference>

Bear in mind that the reference response is only for reference and your main task always is answering the question with the help of the reference answer. \textbf{Do not directly mention the reference answer in your response.}

Now think carefully and give your response to the question:

<start\_of\_question>

<QUESTION\_HERE>

<end\_of\_question>\\
    \bottomrule
    \end{tabular}}
    \caption{Our prompt for rewriting. We use the response from Qwen2.5 as a reference and prompt Gemma-2 to improve the quality of the reference with its own knowledge.}
    \label{tab:prompt}
\end{table*}

We provide our prompt for rewriting in Table \ref{tab:rewriting}. We refer to the prompt in the Mixture-of-Agents (MOA) \cite{wang2024mixture} and provide both question and reference answers for the LLM to rewrite. We find restating the question increases the distribution consistency weight so we mention the question twice. We request the LLM not to mention the reference so that the response can focus on answering the question rather than explaining how the rewritten answer is better than the reference.

\section{Case study of generated samples by InCo-DPO}
\label{sec:example}

\begin{table*}[h]
    \centering
    \scalebox{1}{
    \begin{tabular}{p{0.15\linewidth}p{0.1\linewidth}p{0.6\linewidth}}
        \toprule
        Method&Reward&Content\\
        \midrule
        On-Policy&0.0815&As a large language model, I don't have personal experiences or the ability to interact with the world in a physical way. I can't perform actions in the real world, so I haven't had the opportunity to make a direct, tangible impact on someone's life...\\
        \midrule
        Off-Policy&0.2295&A few years ago, I worked as a volunteer tutor at an after-school program for underprivileged children in my community. One particular student, a 12-year-old girl named Maria, caught my attention...\\
        \midrule
        Continuation&0.2295&\emph{A few years ago}, I was working as a volunteer tutor at a local community center. One day, a young girl named Sarah came to me for help with her math homework. She was struggling with fractions, and her frustration was palpable...\\
        \midrule
        Rewriting&0.2373&A few years ago, while working as a tutor at a community center, I met a young boy named Alex. He was struggling in math, falling behind in his class and visibly losing confidence. He'd come to tutoring sessions quiet and withdrawn, often burying his head in his hands when faced with even simple problems...\\
    \bottomrule
    \end{tabular}}
    \caption{An example of generated samples by InCo-DPO with instruction of \emph{Describe a specific event where you made a significant and positive impact on someone's life}, where \emph{A few years ago} is the prefix from Qwen2.5. Continuation and rewriting receive significantly higher reward compared to on-policy.}
    \label{tab:incodpo_example}
\end{table*}

\begin{table*}[h]
    \centering
    \scalebox{1}{
    \begin{tabular}{p{0.15\linewidth}p{0.1\linewidth}p{0.6\linewidth}}
        \toprule
        Method&Reward&Content\\
        \midrule
        On-Policy&0.1318&It's tough to describe an ion trap visually without a picture, but I can give you a textual representation! Imagine a **small, cylindrical chamber** made of metal, like a tiny metal tube. Inside this chamber, **electric fields** are carefully manipulated to create **confining regions**...\\
        \midrule
        Off-Policy&0.1699&An ion trap is a device used to capture and confine charged particles (ions) using electric and magnetic fields. There are several types of ion traps, each with its own specific design, but they generally share some common features...\\
        \midrule
        Continuation&0.1543&\emph{An ion trap is} a device used to confine charged particles, called ions, using electric and/or magnetic fields. **Here's a simplified description of what it looks like:**...\\
        \midrule
        Rewriting&0.1523&An ion trap is a device that uses electric and magnetic fields to hold charged particles, called ions, in place. Imagine a tiny cage, invisible to the naked eye, built from carefully shaped metal electrodes...\\
    \bottomrule
    \end{tabular}}
    \caption{An example of generated samples by InCo-DPO with instruction of \emph{What does an ion trap look like?}, where \emph{An ion trap is} is the prefix from Qwen2.5. Continuation and rewriting receive significantly higher reward compared to on-policy.}
    \label{tab:incodpo_another_example}
\end{table*}

We provide an example showing how continuation and rewriting can improve the reward compared to on-policy data. As shown in Table \ref{tab:incodpo_example}, for the instruction that requests LLM to provide a specific event, the on-policy data refuses to answer while the prefix \emph{A few years ago} guides the LLM to provide a detailed description of an event, thus improving the reward. Rewriting also improves the reward through prompting the model to answer the instruction with the off-policy response from a strong model as reference. However, the effectiveness of rewriting is heavily dependent on the prompts; when a prompt is provided, the corresponding rewards and weights are predetermined. In contrast, continuation can dynamically adjust the balance between reward values and weights, allowing for better generalization across different models and DPO variants. Moreover, for rewriting, each model requires a specifically tailored rewriting prompt. For instance, a prompt suitable for Gemma-2 is not only ineffective for Llama-3 but may even result in a lower reward compared to the on-policy data after rewriting high-reward off-policy response. Another example can be found in Table \ref{tab:incodpo_another_example}.

\section{Experiment of InCo-DPO on small-scale models}
\label{sec:tiny}
In this section, we verify InCo-DPO on tiny models as they could be essential for on-device deployment. We select SmolLM2-135M-Instruct \cite{allal2024SmolLM2} and Gemma-2-2B-it \cite{team2024gemma} for our experiments, whose parameter is considerably small.

As shown in Table \ref{tab:tiny}, the data synthesized by InCo-DPO outperforms both off-policy and on-policy data, demonstrating its effectiveness even on tiny models. 

\begin{table}[H]
    \centering
    \scalebox{0.65}{
    \begin{tabular}{cccc}
        \toprule
        \textbf{Data} & \multirow{2}{*}{\textbf{Trained Model}} & \multirow{2}{*}{\textbf{External Model}} & \multicolumn{1}{c}{\textbf{AE}} \\ 
        \textbf{Construction} &  &  & \multicolumn{1}{c}{\textbf{LC (\%)}} \\ 
        \midrule
        Off-Policy &SmolLM2-135M-Instruct &Qwen2.5-72B & 0.83 \\ 
        On-Policy &SmolLM2-135M-Instruct &- & 1.02 \\ 
        Ours &SmolLM2-135M-Instruct& Qwen2.5-72B& \textbf{1.32} \\
        \midrule
        Off-Policy &Gemma-2-2B-it &Qwen2.5-72B & 50.67 \\
        On-Policy &Gemma-2-2B-it &-& 50.37 \\
        Ours & Gemma-2-2B-it &Qwen2.5-72B & \textbf{56.08} \\
    \bottomrule
    \end{tabular}}
    \caption{Results of tiny models on different data construction strategies.}
    \label{tab:tiny}
\end{table}

\section{More trials on data construction strategy}
In this section, we discuss other approaches to leveraging off-policy data for constructing preference pairs. In all experiments within this paper, we directly select responses with the maximum and minimum reward values from candidate samples as $y_w$ and $y_l$. Apart from this unconstrained selection method, we also explored a constrained approach for utilizing off-policy responses from SOTA models. Specifically, we forced $y_w$ to be  an off-policy sample and $y_l$ to be an on-policy sample from the model to be trained. We filter out preference pairs if the reward of the off-policy response is lower than that of the on-policy response. However, our experiments reveal that this construction strategy performed poorly, even underperforming the use of purely on-policy data.

We hypothesize that this strategy disrupts the overall preference distribution. Although off-policy responses generally have higher rewards, we observe that on some instructions, on-policy responses achieved higher rewards than off-policy responses. Forcibly rejecting these on-policy responses introduces noise into the preference distribution, leading to suboptimal performance.

\end{document}